\documentclass{ieeeaccess}
\usepackage{cite}
\usepackage{amsmath,amssymb,amsfonts}
\usepackage{graphicx}
\usepackage{textcomp}
\usepackage{algorithm}
\usepackage{listings}
\usepackage{lipsum}
\usepackage{multirow}
\usepackage{lineno,hyperref}
\usepackage{algpseudocode}
\usepackage{booktabs}
\usepackage{mathtools}
\usepackage{upgreek}
\usepackage{soul,color}
\soulregister\cite7
\soulregister\ref7
\soulregister\sc7
\usepackage{caption}
\DeclareCaptionFont{ieeeblue}{\color{accessblue}}
\DeclareCaptionLabelFormat{myformat}{\raggedright\figcapfont{\textbf{#1}\textbf{#2}}}
\def\BibTeX{{\rm B\kern-.05em{\sc i\kern-.025em b}\kern-.08em
T\kern-.1667em\lower.7ex\hbox{E}\kern-.125emX}}

\begin{document}

\history{Received April 16, 2021, accepted May 10, 2021, date of publication May 17, 2021, date of current version May 28, 2021.}
\doi{10.1109/ACCESS.2021.3080837}

\title{Entertainment chatbot for the digital inclusion of elderly people without abstraction capabilities}
\author{\uppercase{Silvia García-Méndez}\authorrefmark{1},\uppercase{ Francisco de Arriba-Pérez}\authorrefmark{1},\uppercase{ Francisco J. Gonz\'alez-Casta\~no\authorrefmark{1},\uppercase{ José A. Regueiro-Janeiro}\authorrefmark{1},\uppercase{ and Felipe Gil-Casti\~neira}\authorrefmark{1}}}
\address[1]{Information Technologies Group, atlanTTic, University of Vigo, Telecommunication Engineering School, Campus, 36310 Vigo, Spain}
\tfootnote{This work was supported by {\sc mineco} grant {\sc tec} {\sc 2016-76465-C2-2-R} and Xunta de Galicia grant {\sc grc} 2018/053, Spain. The authors are indebted to Asociaci\'on de Familiares de enfermos de Alzheimer y otras demencias de Galicia ({\sc afaga}) for providing us with gerontology expertise and valuable design recommendations.}

\markboth
{Silvia García-Méndez \headeretal: Entertainment chatbot for the digital inclusion of elderly people without abstraction capabilities}
{Silvia García-Méndez \headeretal: Entertainment chatbot for the digital inclusion of elderly people without abstraction capabilities}

\corresp{Corresponding author: Silvia García-Méndez (e-mail: sgarcia@gti.uvigo.es).}

\begin{abstract}
Current language processing technologies allow the creation of conversational chatbot platforms. Even though artificial intelligence is still too immature to support satisfactory user experience in many mass market domains, conversational interfaces have found their way into {\it ad hoc} applications such as call centres and online shopping assistants. However, they have not been applied so far to social inclusion of elderly people, who are particularly vulnerable to the digital divide. Many of them relieve their loneliness with traditional media such as TV and radio, which are known to create a feeling of companionship. In this paper we present the {\sc eber} chatbot, designed to reduce the digital gap for the elderly. {\sc eber} reads news in the background and adapts its responses to the user's mood. Its novelty lies in the concept of ``intelligent radio'', according to which, instead of simplifying a digital information system to make it accessible to the elderly, a traditional channel they find familiar -background news- is augmented with interactions via voice dialogues. We make it possible by combining Artificial Intelligence Modelling Language, automatic Natural Language Generation and Sentiment Analysis. The system allows accessing digital content of interest by combining words extracted from user answers to chatbot questions with keywords extracted from the news items. This approach permits defining metrics of the abstraction capabilities of the users depending on a spatial representation of the word space. To prove the suitability of the proposed solution we present results of real experiments conducted with elderly people that provided valuable insights. Our approach was considered satisfactory during the tests and improved the information search capabilities of the participants.
\end{abstract}


\titlepgskip=-15pt

\maketitle

\section{Introduction}
\label{introduction}

Humans are innately social and need companionship \cite{Caligiuri2002,Blaakilde2018}. Social inclusion of elderly people is essential to improve their quality of life in our ageing society. 
According to the Spanish National Statistics Institute, Spain had ageing indexes of 118.26\% in 2017 and 120.46\% in 2018. The population in their sixties grew to 19.20\% in 2018 and 54.19\% of these needed care\footnote{Available at {\tt https://www.ine.es/en/welcome.shtml}, March 2021.}. 

The rapid expansion of digital technologies has had appalling social benefits. Few will deny that these have been unbalanced across different population groups. In the case of elderly people, several works have noted their low level of engagement due to the sometimes insuperable digital gap \cite{Czaja2007,Favela2015,Fleming2018}. Alleviating their digital exclusion in a way they find acceptable is the main motivation of this work.

The digital gap impedes many elderly people to access digital content and even technology-aware people eventually experience this gap when their cognitive capabilities fade. Commercial conversational systems like Siri, Google Assistant and Cortana simplify access to digital media, but they require technological skills often beyond the reach of the senior population \cite{Kleinberger2007}. This population typically alleviates solitude by continuously consuming traditional broadcast media such as television and radio \cite{Ostlund2010}. We have taken this behaviour into account in the design.

We are interested in user-centric approaches to close this digital gap. More specifically, this work involves the development of a user-friendly assistant to make Internet media accessible to elderly people. The idea we are pursuing is to create an assistant that will be perceived as an ``intelligent radio" companion.

Chatbots appear to be a convenient solution. A chatbot, or chatting robot, is a system that enables human-to-computer interaction through text or natural language ({\sc nl}) \cite{Mahapatra2012}. Ideally it should sustain human-like conversations, but this is clearly beyond the current state-of-the-art \cite{Skjuve2019HelpInteraction}, even for chatbots based on Artificial Intelligence ({\sc ai}) technologies such as Natural Language Generation ({\sc nlg}).

In general, chatbots have three main components: a user interface, a normaliser and a knowledge base \cite{A.2015,Liu2018}. Some famous chatbots are {\sc alice} \cite{AbuShawar2015}, SimSimi\footnote{Available at {\tt https://rebot.me/luisse-kate-barcelona}, March 2021.} and Cleverbot\footnote{Available at {\tt https://www.cleverbot.com}, March 2021.}. A common approach to chatbot development is to preprogram questions and expressions to respond to user input in the form of questions, phrases and words \cite{Gupta2015}. Additional content is automatically added by building knowledge on the conversations held.

Given the expected behaviour of the target users (who find radio and TV entertaining), we do not need the chatbot to be as intelligent and empathetic as a human. The idea is to intersperse the news items that the chatbot reads with light dialogues that act as news ``connectors'', fostering engagement without demanding too much attention from the user. In brief, the chatbot serves as an intelligent radio capable of reading news and holding a limited dialogue. The chatbot applies Sentiment Analysis ({\sc sa}) to user responses to adapt the dialogue flow to the user's mood. This increases the value of chatbot-induced dialogues to gather information on user preferences. Using Google distance measurements \cite{Cilibrasi2007}, we demonstrate that even unfocused information extracted from these dialogues help conduct personalised information searches on the web in a manner that is transparent to end users. 

Summing up, from a conceptual point of view, we present an ``intelligent radio" by augmenting an information channel the elderly find familiar -background news- with interactions by voice dialogues. The system allows accessing digital content of interest, by combining words extracted from user answers to chatbot questions with keywords extracted from the news items. This permits defining metrics of the abstraction capabilities of the users depending on a spatial representation of the word space.

The rest of this paper is organised as follows. Section \ref{related_work} discusses related work. Section \ref{nlp} presents the proposed solution. Section \ref{evaluation} reports experimental tests with real users. Finally, Section \ref{conclusions} concludes the paper.

\section{Related Work}
\label{related_work}

Recent {\sc ai} advances have enabled chatbots to automatically generate responses to user requests \cite{Xu2017}. As a result, they are now widely used in domains such as business \cite{Thomas2016}, education \cite{Robins2005RoboticSkills,Carlander-Reuterfelt2020} and healthcare \cite{Kumar2016,Ren2020}. However, they are still very far from mimicking human intelligence \cite{Skjuve2019HelpInteraction}. Given the current state-of-the-art, it is broadly agreed that chatbots should be oriented to highly focused tasks in specific application domains and that their interactions should be precise and short, as people find easier to communicate and engage for longer with short-message dialogue systems \cite{Hill2015}. 

Chatbots are based on retrieval or generative models. Retrieval models apply predefined rules according to a rigid syntax. One popular technology for the rapid development of chatbot personalities is the Artificial Intelligence Markup Language ({\sc aiml}) \cite{Galvao2004, Thomas2016}. Generative models, based on advanced deep learning models \cite{Su2018} combined with Natural Language Programming ({\sc nlp}) \cite{Baby2018,Oh2017,Wen2015}, are more complex. Optimal approaches vary depending on the application domain and the demographics of the target users. From a syntactic point of view, {\sc aiml} is almost error-free, but {\sc nlp} techniques allow for more human-like {\sc nl} dialogue flows. We decided to combine these two approaches to exploit their respective advantages. 

The application in this paper is related to healthcare \cite{Su2018} and entertainment \cite{Noh2017}, two domains in which chatbot applications are still at an early stage. Online platforms for general medical care such as HealthTap\footnote{Available at {\tt https://www.healthtap.com}, March 2021.}, Oscar\footnote{Available at {\tt https://www.hioscar.com}, March 2021.} and Molly\footnote{Available at {\tt http://www.sensely.com}, March 2021.} support interactions between patients and doctors. In \cite{Kenny2008} a solution for mental health diagnosis and clinician training on virtual patients was described, and in \cite{Yasuda2014} an assistant chatbot for people with dementia was presented. The focus of chatbot applications for elderly people is typically therapeutic. 
One of the few exceptions was the personal assistant in the RobAlz project \cite{Salichs2016} that also reads stories and plays games with people with early-stage dementia. However, its dialogues may be considered canned-text since the chatbot employs predefined clauses and only invites users to continue talking. Conversely, we aim at providing {\it ad hoc} dialogues within the context of the news not only to entertain but also to increase the feeling of companionship of the ``digital radio''.

Other entertainment chatbots \cite{Johnson2016,Correia2016,Aaltonen2017} like Edu\-Robot \cite{Cahyani2017} have singing and storytelling capabilities but they have not been designed for an audience of elderly people. On the contrary, we have actively considered these target users into the design process thanks to the collaboration with Asociaci\'on de Familiares de enfermos de Alzheimer y otras demencias de Galicia ({\sc afaga}\footnote{Available at {\tt https://afaga.com}, March 2021.}) which provided us with gerontology expertise and valuable recommendations.

Even though newscaster chatbots are available \cite{Matsumoto2007JournalistWorld}, they have not been provided with empathy so far. Note that we are only considering chatbots. In the case of robots there has been plenty of research on empathetic capabilities \cite{Augello2008,Oh2017} and even commercial products are available. Sony's {\sc aibo} robotic dog \cite{Pransky2001}, for example, could express happiness, anger, fear, sadness, surprise and dislike by means of tail and body movements and the colour and shape of its eyes. That is, robots typically express empathy by simulating pets, instead of by adapting generated language to user's expressions as in this work.

Considering the state-of-the-art in different research areas (health, entertainment, etc.), the main conceptual contribution of this work is a chatbot that has been realistically conceived from the onset for inclusive elderly entertainment through information delivery. It alternates newscasts with light dialogues about the news items that are adapted to the user's mood. Descriptions of news items are automatically extracted from these dialogues for content recommendation purposes. As far as we know, the extraction of search information from dialogues with elderly people to improve their abstraction capabilities --as shown in Section \ref{abstraction_measurement}-- has not been previously studied. We thus leverage state-of-the-art {\sc nlg} and {\sc sa} technologies to engage the user and reduce the digital divide. To prevent frustration the chatbot must provide flexible responses instead of repeatedly resorting to clauses such as `Sorry, I didn't understand you' \cite{Jenkins2007}. Some degree of chatbot personality is crucial for motivating user engagement \cite{Andrews2012}. Here is where {\sc sa} and {\sc nlg} are essential. 

Regarding these underlying key technologies, {\sc nlg} allows creating complete, coherent, human-like texts \cite{ReiterEhudandDale2000} from linguistic information (words, sentences, texts) and non-linguistic information (objective and visual data). {\sc nlg} applications include automatic question generation \cite{Wang2009ComputerRecognition}, summaries \cite{Genest2011} and text simplification \cite{Rennes2015}, and their possibilities have multiplied thanks to Machine Learning. In particular, automatic text generation (whose only required inputs are meaningful words such as nouns, verbs and adjectives) has attracted great attention \cite{Garcia-Mendez2019AExpansion}.

{\sc sa} \cite{Cambria2016} is a hot field in social media research \cite{Nguyen2015,Poecze2018,Sohrabi2019}. It allows inferring the polarity of a text (positive, negative or neutral). Among its many potential business applications \cite{Kirilenko2018,Vyas2019}, it has been used to analyse consumer opinions in online media \cite{Chaurasiya2019}, politics \cite{Martin-Gutierrez2018} and sociology \cite{Eliacik2018}. It can work at document \cite{Chaurasiya2019}, sentence \cite{Eliacik2018} and even entity/aspect level \cite{Ma2018}.

\section{Solution Design}
\label{nlp}

The solution in this work is designed to entertain the elderly by leveraging the latest advances in different technologies. {\sc sa} is the key building block for achieving the limited but key empathetic capabilities we need for the dialogue stages between newscasts. {\sc nlg} allows varied, flexible and naturally sounding interactions with the user. To properly match keywords extracted from user speech and to keep conversation flowing and engage the user in an entertaining exchange, the design considered personality, conversation flow, empathy/emotion, chatbot dialogue structure, graphics and accessibility. Regarding the latter, it is based on facial recognition instead of keywords, so that the system starts the dialogue when it detects that the user is present.

\subsection{News Service}
\label{news}
We chose a news service as the basis of the application because listening to the news, especially local ones, is a main pastime of elderly people. As previously mentioned, traditional entertainment media such as television and radio have a companionship effect \cite{Ostlund2010}.

We selected a varied range of topics that might be of interest to the target population: accessibility, environment, health, leisure, public services, retirement, social services, sport, 
technology and transport.

At the end, the news service allows the {\sc eber} bot engaging elderly people thanks to the varied and updated news presented to them, preventing monotony with continuous changes of dialogue topics. We remark that the dialogues are short and that their purpose is to ``connect'' news while, at the same time, obtaining user opinions and preferences. The system automatically gathers the news items from the Application Programming Interface ({\sc api}) of the Spanish National Radio and Television (RTVE\footnote{Available at {\tt https://www.rtve.es/api}, April 2021.}) channel with {\sc get} queries, using the {\tt tematicas} and {\tt noticias} services of the {\sc api}, so that it is possible to extract news about certain topics.

\subsection{NLG Module}
\label{nlg_module}

\begin{figure*}[!htbp]
 \centering
 \includegraphics[scale=0.18]{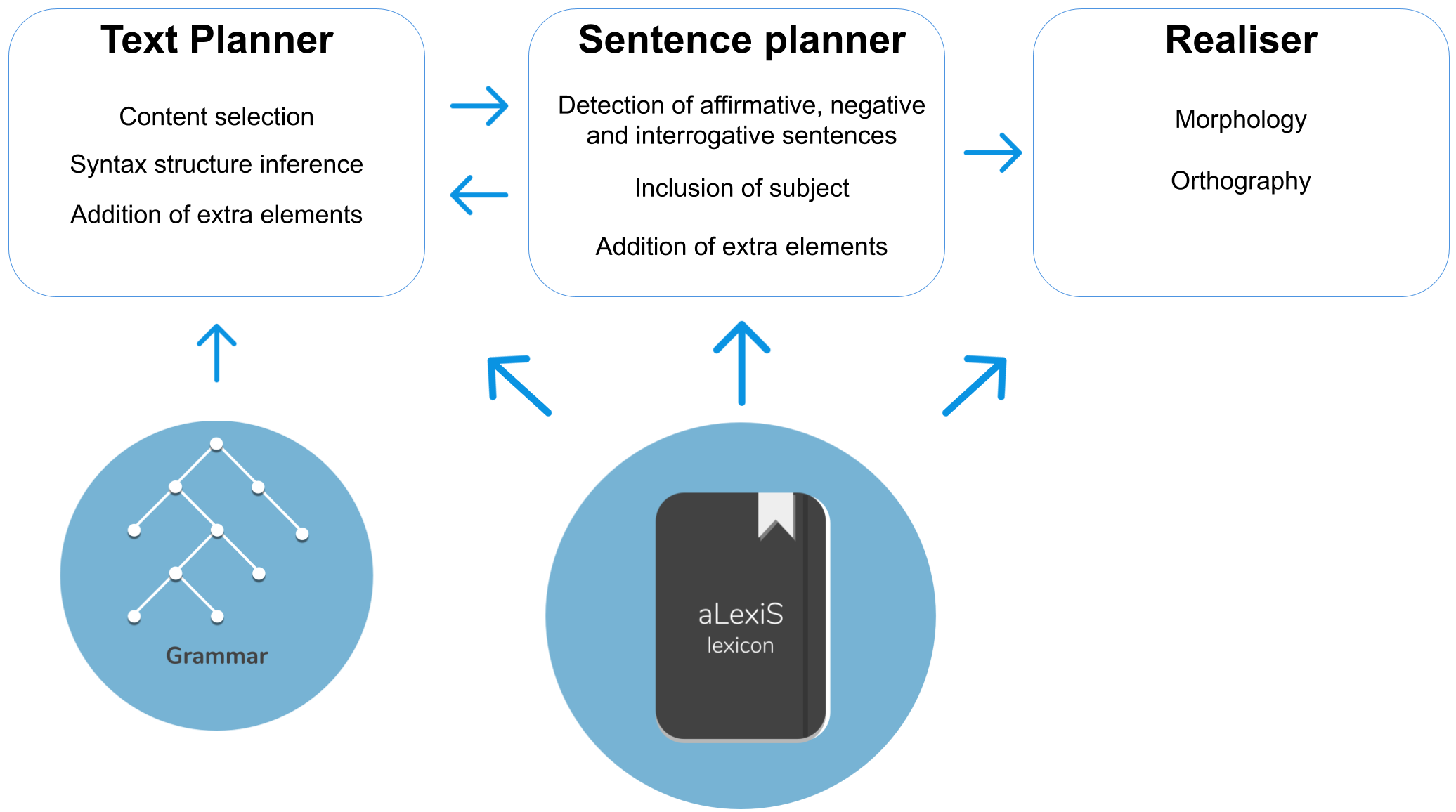}
 \caption{Tree-stage NLG module.}
 \label{fig:nlg}
\end{figure*}

Figure \ref{fig:nlg} shows the three-stage {\sc nlg} module. It is based on our automatic text-to-text Spanish generator using keywords as input words \cite{Garcia-Mendez2018,Garcia-Mendez2019}.

The first stage, the \textit{Text Planner}, comprises content selection and text structuring tasks. Input data are meaningful words for creating an {\sc nl} sentence in Spanish. The stage infers the best text structure from the information in the grammar. The grammar considers nominal and coordinated nominal, adjectival, adverbial and prepositional phrases.

The second \textit{Sentence Planner} stage deals with lexicalisation, which consists of adding extra words automatically (once the syntactic structure has been inferred, it may be necessary to include determiners, prepositions and conjunctions). This is the reason for the feedback between the second and first stages in Figure \ref{fig:nlg}.

The third stage, the \textit{Realiser}, conducts morphological inflections to create a coherent and grammatically correct sentence in Spanish. 

Thus, the main {\sc nlg} actions in Figure \ref{fig:nlg} are:

\begin{itemize}

\item Syntax structure inference (\textit{Text Planner}).

\item Detection of affirmative, negative and interrogative sentences (\textit{Sentence Planner}).

\item Inclusion of subject (\textit{Sentence Planner}).

\item Addition of extra elements (\textit{Text Planner} and \textit{Sentence Planner}).

\item Morphological inflections (\textit{Realiser}).

\item Orthographic rules (\textit{Realiser}).

\end{itemize}

These actions take linguistic knowledge from our \textit{aLexiS} lexicon \cite{Garcia-Mendez2018,Garcia-Mendez2019} to adjust the sentence features (number, gender, person, tense).
Specifically, the {\sc nlg} module creates affirmative and negative clauses using keywords (verbs, for example), and adjective and noun phrases extracted from news content and user utterances. For that purpose, we use the spaCy\footnote{Available at {\tt https://spacy.io}, March 2021.} parser. At the end, it creates complete, coherent and correct natural language sentences by applying the aforementioned {\sc nlg} actions.

To illustrate the operation of the module, let us consider a complete example of text expansion. Assuming the user says \textit{Me parece una noticia interesante} `I think this news item seems interesting', the chatbot's response is \textit{Yo considero que no parece interesante} `I consider it does not seem interesting'. To create this reply, which is closely related to the previous user utterance, firstly the {\sc nlg} module extracts the subject from the user clause \textit{yo} `I' and the main verb \textit{parece} `seem' (this linguistic information is taken from \textit{aLexiS}). Thus, for short dialogues that are highly related to the news, the module can infer from the verbs that the users are expressing opinions. Secondly, the module extracts the adjective from the response of the user, \textit{interesante} `interesting'. 

\subsection{SA Module}
\label{sa_module}

\begin{figure*}[!htbp]
 \centering
 \includegraphics[scale=0.2]{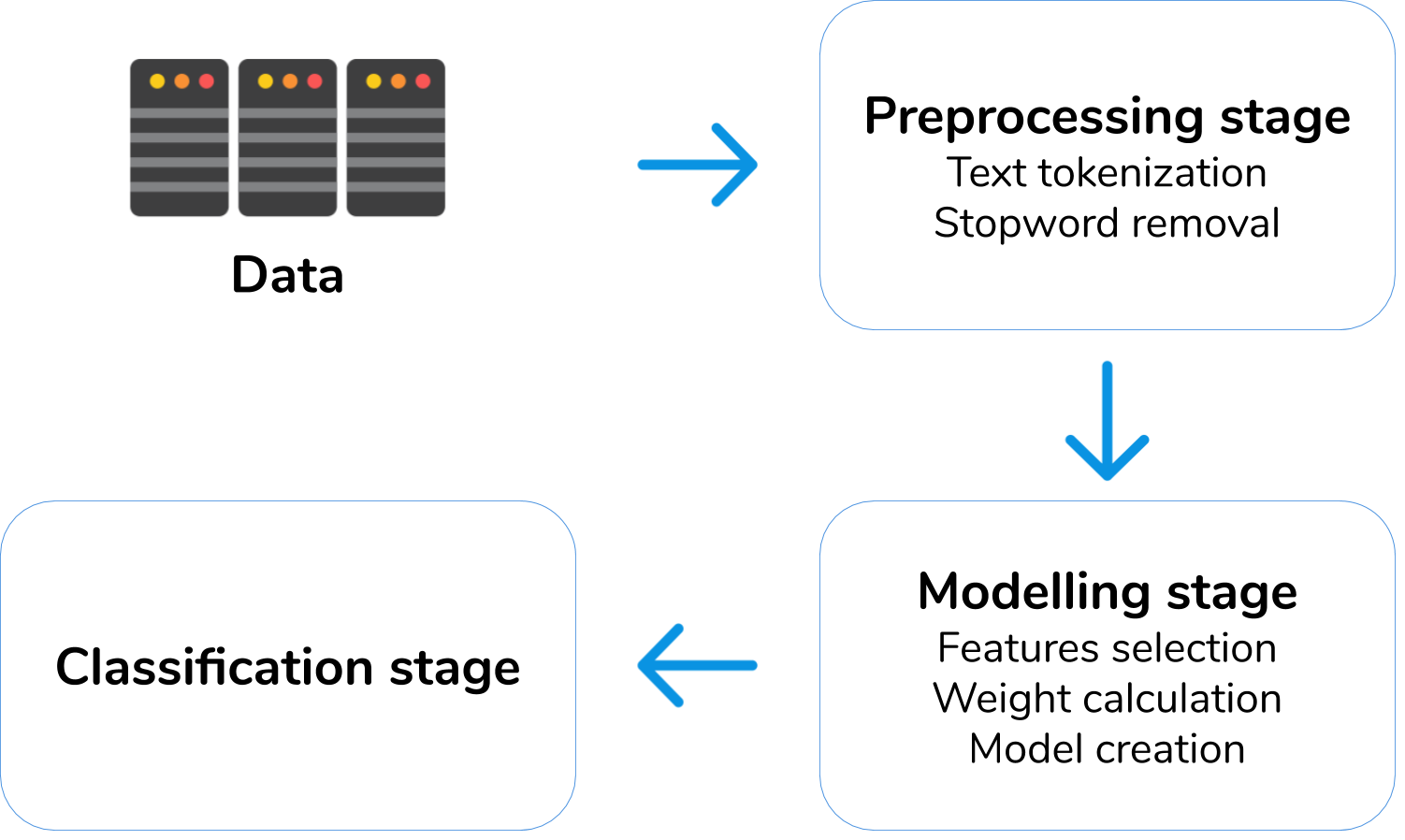}
 \caption{Flow diagram of the {\sc sa} module.}
 \label{sadiagram}
\end{figure*}

We have designed a three-stage {\sc sa} module (Figure \ref{sadiagram}) to provide {\sc eber} with empathetic capabilities. It classifies short texts into three polarity levels (negative, neutral and positive). In order to evaluate the effectiveness of the module, we trained some selected Machine Learning algorithms from the Scikit-Learn Python library\footnote{Available at {\tt https://scikit-learn.org/stable/
supervised\_learning.html\#supervised-learning}, April 2021.}, Gradient Descent, Decision Tree and Random Forest, on a dataset of 676 negative, 1,334 neutral and 918 positive clauses. Table \ref{tab:features} shows the features we considered: character and word $n$-grams \cite{Brants2007} and three word dictionaries of deniers (adverbs with negative meaning), positive and negative words.

\begin{table*}[ht!]
\centering
\small
\caption{\label{tab:features} Input features in the Machine Learning model.}
\begin{tabular}{lll}
\hline
\bf Type & \bf Feature name & \bf Description \\\toprule

Textual & CHAR\_GRAMS & \begin{tabular}[c]{@{}p{8.5cm}@{}} Sequence of character $n$-grams\end{tabular}\\

 & WORD\_GRAMS & \begin{tabular}[c]{@{}p{8.5cm}@{}} Sequence of word $n$-grams\end{tabular}\\ \hline

Numerical & NEG\_ADVERBS & \begin{tabular}[c]{@{}p{8.5cm}@{}} Amount of adverbs with negative meaning\end{tabular}\\
 
 & POS\_WORDS & \begin{tabular}[c]{@{}p{8.5cm}@{}} Amount of words with positive meaning\end{tabular}\\
 
 & NEG\_WORDS & \begin{tabular}[c]{@{}p{8.5cm}@{}} Amount of words with negative meaning\end{tabular}\\
\bottomrule
\end{tabular}
\end{table*}

\begin{table*}[!htbp]
\centering
\caption{\label{tab:resultsnumericaltest1} Evaluation of Machine Learning algorithms for the SA-module.}
\small
\begin{tabular}{cccc}
\toprule
\bf Classifier & \boldmath \bf $P_{\mbox{\footnotesize macro}}$ & \bf \boldmath $F_{\mbox{\footnotesize macro}}$ & \bf Accuracy \\ \hline
Gradient Descendent & 76.20\% & 74.11\% & 74.52\% \\
Decision Tree & \bf 80.65\% & \bf 79.71\% & \bf 79.99\% \\
Random Forest & 74.82\% & 72.97\% & 73.49\% \\
\bottomrule
\end{tabular}
\end{table*}

\begin{figure*}[!htbp]
 \centering
 \includegraphics[scale=0.36]{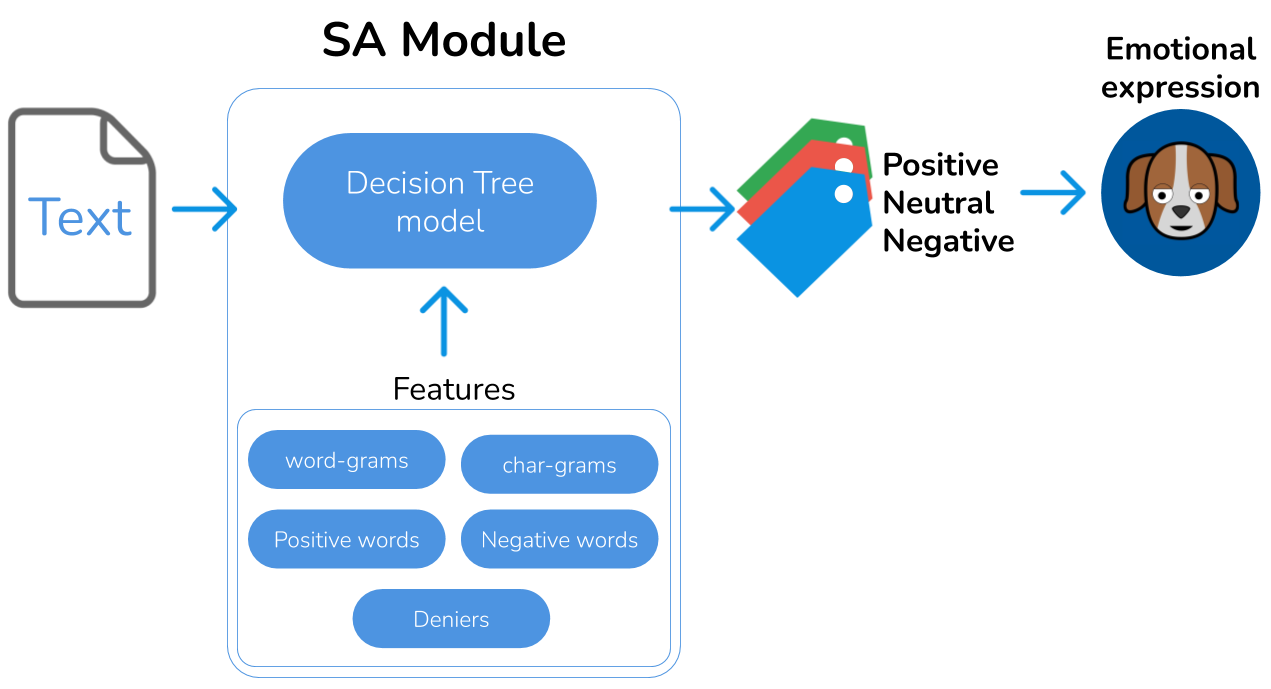}
 \caption{Decision Tree classifier scheme.}
 \label{dtstage}
\end{figure*}

Then, we applied an attribute selector to extract the most relevant features. Specifically, we employed the SelectPercentile\footnote{Available at {\tt https://scikit-learn.org/stable/modules
/feature\_selection.html}, April 2021.} method from the Scikit-Learn Python library, as it outperformed other alternatives (SelectFromModel, SelectKBest, and RFECV). This method selects features according to a highest score percentile. We set a $\chi^2$ score function and an 80th percentile threshold. Table \ref{tab:resultsnumericaltest1} presents the performance of the Machine Learning algorithms. The Decision Tree classifier was the best alternative for the {\sc sa} module. It attained 79.99\% accuracy and 79.71\% $F_{\mbox{\footnotesize macro}}$ with 10-fold cross validation on class-balanced testing subsets. Figure \ref{dtstage} shows the scheme of this classifier. As previously said, the conversational system employs the output of the {\sc sa} module to adjust the facial expression of its avatar to the user mood. Following the example in Section \ref{nlg_module} in which the user expresses an opinion, the {\sc sa} module detects that the clause \textit{Me parece una noticia interesante} has positive polarity. Consequently, the avatar shows a happy face.

The {\sc nlg} module employs {\sc sa} knowledge to avoid monotony by adjusting the polarity of the dialogue depending on the polarity of user responses. With a configurable probability, they take opposite values, as in the example in Section \ref{nlg_module}. We remark that the dialogue is very light and that it is only intended to engage the users and obtain their opinion on the news. When the {\sc nlg} module is unable to extract keywords or phrases to generate coherent empathetic text, it selects positive or negative sentences from a set of predefined templates based on the knowledge of the {\sc sa} module.

\subsection{Chatbot Design}
\label{methodology}

We seek to assist elderly people with limited information abstraction capabilities, to help them find information of their interest (although this research would also be valid for other target groups with similar limitations). Traditional broadcast entertainment media has a proved ``companionship'' effect for the elderly \cite{Ostlund2010}. This is why we focused on news complemented with short dialogues as content delivery format.

In the following sections we elaborate on the different modules of the {\sc eber} chatbot, which has been designed to reduce the digital gap for those target users. Accordingly, we have actively considered them into the design process.

Apart from well-known accessibility standards, first and foremost we have followed the advice of {\sc afaga}. They have provided us with valuable design recommendations from their Cognitive Stimulation Program through New Technologies\footnote{Available at {\tt https://afaga.com/es/servicios/taller-de
-nuevas-tecnologias}, April 2021.}, an innovative experience to promote new technologies in devices for the elderly.

\subsubsection{{\sc aiml} Chatbot Personality}
 
Since the purpose of the chatbot is well defined (to obtain opinions about news from elderly people through light conversations), we decided to create our own {\sc aiml}-based knowledge base rather than reuse an existing chatbot. We leveraged {\sc aiml} features such as \textit{star}, \textit{srai} and \textit{random} as well as pattern extraction with keywords. This allows the chatbot starting the conversation with a controlled dialogue flow, using short questions about daily routines and mood. These questions are extracted from predefined templates as described in sections \ref{con_decision} and \ref{sec_dialogue}.

\subsubsection{Conversation Decisions}
\label{con_decision}

Table \ref{default} shows some examples of default answers to keep conversation flowing and prevent frustration. 
When the {\sc eber} chatbot detects that the user cannot understand something, it asks him/her to explain why. By doing so the conversation feels more natural and the user feels more involved.

\begin{table*}[!htbp]
\centering
\caption{\label{default}Default answers and respective chatbot connectors.}
\begin{tabular}{cc}
\hline
\bf Option & \bf Default answers \\\toprule
1 & \multicolumn{1}{l}{{\em Eres una persona muy interesante} `You're a really interesting person'}\\
2 & \multicolumn{1}{l}{{\em Me gustar\'ia conocerte m\'as} `I'd like to get to know you better'}\\
3 & \multicolumn{1}{l}{{\em Me gustar\'ia saber m\'as de ti} `I'd like to know more about you'}\\
4 & \multicolumn{1}{l}{{\em Cu\'entame m\'as} `Tell me more'}\\\bottomrule

\bf Option & \bf Connector \\\toprule

\multirow{4}{*}{1}
& \multicolumn{1}{l}{{\em ?`Usas mucho el tel\'efono m\'ovil?} `Do you use your cell phone a lot?'}\\
& \multicolumn{1}{l}{{\em ?`Tuviste alguna mascota?} `Have you ever had a pet?'}\\
& \multicolumn{1}{l}{{\em ?`Cu\'al fue tu primer empleo?} `Which was your first job?'}\\
& \multicolumn{1}{l}{[...]}\\\hline

\multirow{4}{*}{2}
& \multicolumn{1}{l}{{\em ?`Cu\'al es tu color favorito?} `What's your favourite colour?'}\\
& \multicolumn{1}{l}{{\em ?`Cu\'al es tu deporte favorito?} `What's your favourite sport?'}\\
& \multicolumn{1}{l}{{\em ?`Te gusta el f\'utbol?} `Do you like football?'}\\
& \multicolumn{1}{l}{[...]}\\\hline

\multirow{4}{*}{3}
& \multicolumn{1}{l}{{\em ?`Te gusta hacer deporte?} `Do you like practising sport?'}\\
& \multicolumn{1}{l}{{\em ?`Te gusta ir de compras?} `Do you like shopping?'}\\
& \multicolumn{1}{l}{{\em ?`Te gusta jugar a las cartas?} `Do you like playing cards?'}\\
& \multicolumn{1}{l}{[...]}\\\hline

\multirow{4}{*}{4}
& \multicolumn{1}{l}{{\em ?`Te gusta leer?} `Do you like reading?'}\\
& \multicolumn{1}{l}{{\em ?`A qu\'e jugabas de peque\~no?} `Which games did you play when you were a child?'}\\
& \multicolumn{1}{l}{[...]}\\\bottomrule
\end{tabular}
\end{table*}

Precise, short chatbot interactions are adequate for our application domain. They are also beneficial for news recommendation purposes, since they reduce meaningless concepts in the dialogues and allow the conversation to move from general to more specific topics. 

\subsubsection{Accessibility and Graphics}
\label{accGrap}

In this regard, the system is simply activated through facial recognition (using the OpenCV library\footnote{Available at {\tt https://opencv.org}, April 2021.} trained with an eye-detection dataset\footnote{Available at {\tt https://github.com/opencv/opencv/blob/
master/data/lbpcascades/lbpcascade\_frontalface.xml}, April 2021.}) or voice commands. All user interactions with the system are by voice, which is considered the most natural access interface for the elderly \cite{Liu2018} and thus fosters a more engaging conversation experience \cite{Pereira2019}. For this purpose we transform Spanish speech into text and vice versa, as input and output of the conversational system, with the Google Voice Android Software Development Kit ({\sc sdk}) library\footnote{Available at {\tt https://developer.android.com/reference/
android/speech/SpeechRecognizer}, April 2021.}.

Furthermore, we keep the graphic appearance of the chatbot as simple as possible by following accessibility conventions for people with vision and hearing impairments \cite{Diaz-Bossini2014,FaisalMohamedYusof2014} (full screen mode, colour contrast, capitalisation, font and icon size, volume configuration, etc.). Additionally, we use visual references to guide the user through the dialogue stage. For example, since previous research has proved the benefits of using pets as emotional avatars for the elderly \cite{Sharkey2012}, the {\sc eber} chatbot interface is an animated dog, and the system presents to the user the lead paragraph of a news item as a summary.

\subsubsection{Empathy and Emotion}

The {\sc sa} module allows guessing the user mood from his/her answers and provides feedback to the conversational system. Based on {\sc sa} knowledge, the {\sc nlg} module presents questions to the user that are aligned with his/her opinion or refute it with configurable probability. Furthermore, the system adjusts the facial expression of its avatar according to the user mood. For example, the dog's face will become sad if {\sc eber} detects a negative emotion.

\subsubsection{Dialogue Structure}
\label{sec_dialogue}

During the dialogue stage, we keep the interactions of the system precise and short due to the fact that people, and especially the elderly, find short dialogue systems simpler and keep engaged for longer \cite{Hill2015}.

Ontologies and similar models only yield good dialogue structuring results when the expected variation is minimal. In the target scenario, however, users are free to express any ideas about the news they are presented with. To handle this disparity we combined {\sc aiml}, {\sc nlg} and {\sc sa}. Dialogues are structured into two parts: {\sc aiml}-based interactions and {\sc nlg}-based replies. When the {\sc eber} chatbot starts a dialogue, it randomly selects a clause for the first interaction (see sample of options in Table \ref{dialogue_flow}). User responses in turn are expected to contain words such as \textit{bien} `well' (positive answers) or \textit{mal} `bad' (negative answers) (see interaction 2 in the example in Table \ref{dialogue_flow}). The chatbot answer depends on the polarity of the user's answer in the previous interaction (see interaction 3 in Table \ref{dialogue_flow}). The chatbot completes its answer with a dialogue connector (see examples in Table \ref{connectors}).

\begin{figure*}[!htbp]
\centering
\includegraphics[scale=0.1]{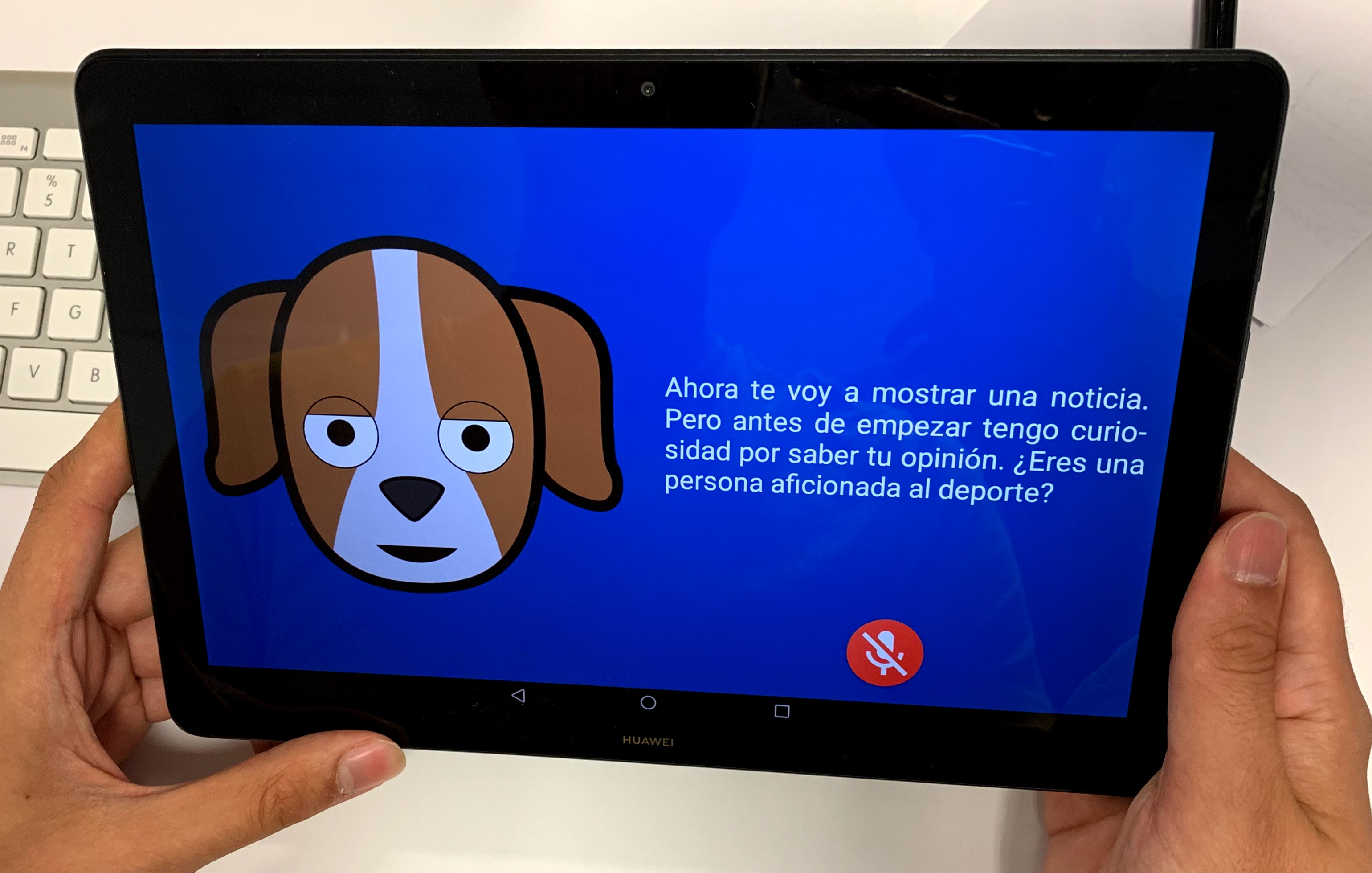}
\caption{\label{eberbot} {\sc eber} chatbot interface. The muted microphone indicates that it is {\sc eber}'s turn to speak.}
\end{figure*}

\begin{table*}[!htbp]
\centering
\caption{\label{dialogue_flow}Dialogue flow interactions, sample.}
\begin{tabular}{ccc}
\hline
\bf Int. \# & \bf Speaker & \bf Options \\\toprule
\multirow{6}{*}{1} & \multirow{6}{*}{Chatbot}
& \multicolumn{1}{l}{{\em ?`Has dormido bien?} `Did you sleep well?'}\\
& & \multicolumn{1}{l}{{\em ?`C\'omo te encuentras hoy?} `How are you today?'}\\
& & \multicolumn{1}{l}{{\em ?`C\'omo te sientes hoy?} `How are you feeling today?'}\\
& & \multicolumn{1}{l}{{\em ?`Qu\'e tal te encuentras?} `How are you doing?'}\\ 
& & \multicolumn{1}{l}{{\em ?`Qu\'e tal est\'as?} `How are you?'}\\
& & \multicolumn{1}{l}{{\em ?`Qu\'e tal est\'a tu familia?} `How's your family?'}\\\hline

\multirow{8}{*}{2} & \multirow{8}{*}{User}
& \multicolumn{1}{l}{{\em $\wedge$ bien} `well' $\wedge$}\\
& & \multicolumn{1}{l}{{\em $\wedge$ genial} `great' $\wedge$}\\
& & \multicolumn{1}{l}{{\em $\wedge$ fant\'astico} `awesome' $\wedge$}\\
& & \multicolumn{1}{l}{{\em $\wedge$ fant\'astica} `awesome' $\wedge$}\\\cline{3-3}
& & \multicolumn{1}{l}{{\em $\wedge$ mal} `bad' $\wedge$}\\
& & \multicolumn{1}{l}{{\em $\wedge$ no muy bien} `not very well' $\wedge$}\\
& & \multicolumn{1}{l}{{\em $\wedge$ fatal} `really bad' $\wedge$}\\
& & \multicolumn{1}{l}{{\em $\wedge$ horrible} `terrible' $\wedge$}\\\hline

\multirow{8}{*}{3} & \multirow{8}{*}{Chatbot}
& \multicolumn{1}{l}{{\em Me alegro por ti} `I'm happy for you'}\\
& & \multicolumn{1}{l}{{\em !`Eso es genial!} `That's great!'}\\
& & \multicolumn{1}{l}{{\em !`Eso est\'a muy bien!} `That's brilliant!'}\\\cline{3-3}
& & \multicolumn{1}{l}{{\em Bueno, poco a poco} `OK, step by step'}\\
& & \multicolumn{1}{l}{{\em Espero hacerte sentir mejor} `I hope I'll make you feel better'}\\
& & \multicolumn{1}{l}{{\em !`\'Animo!} `Cheer up!'}\\\bottomrule
\end{tabular}
\end{table*}

Before reading a news item, the chatbot announces its intention by stating \textit{Ahora te voy a mostrar una noticia, pero antes de empezar...} `Now I'm going to read some news to you, but before we start...' followed by a sentence like:
\begin{itemize}
 \item \textit{Tengo una duda} `I have a question'
 \item \textit{Tengo curiosidad por saber tu opini\'on } `I'm curious about your opinion'
\end{itemize}

Once the chatbot receives the user's answer it says \textit{Vale, vamos entonces con noticias relacionadas con esta tem\'atica} `OK, then let's listen to some news on this topic' and reads the news item.

\begin{table*}[!htbp]
\centering
\caption{\label{connectors}Chatbot connectors, example.}
\begin{tabular}{c}
\hline
\bf Connectors \\\toprule
\multicolumn{1}{l}{{\em H\'ablame de ti} `Tell me about yourself'}\\
\multicolumn{1}{l}{{\em Cu\'entame cosas sobre ti} `Tell me something about you'}\\\hline
\multicolumn{1}{l}{{\em ?`Qu\'e tienes pensado hacer hoy?} `What plans have you got today?'}\\
\multicolumn{1}{l}{{\em ?`Dar\'as un paseo hoy?} `Are you going to go for a walk today?'}\\
\multicolumn{1}{l}{{\em ?`Has quedado con alg\'un amigo?} `Are you meeting a friend?'}\\
\multicolumn{1}{l}{{\em ?`Has quedado con alguien?} `Are you meeting anybody?'}\\ 
\multicolumn{1}{l}{{\em ?`Qu\'e har\'as el fin de semana?} `What are you going to do at the weekend?'}\\
\multicolumn{1}{l}{{\em ?`A d\'onde tienes pensado ir hoy?} `Where are you going today?'}\\
\multicolumn{1}{l}{{\em ?`De qu\'e trat\'o el \'ultimo sue\~no que tuviste?} `What was your last dream about?'}\\
\multicolumn{1}{l}{{\em ?`En qu\'e est\'as pensando ahora mismo?} `What are you thinking right now?'}\\
\bottomrule
\end{tabular}
\end{table*}

Finally, the {\sc nlg} and {\sc sa} modules support the exchange between the chatbot and the user, providing {\sc eber} with valuable information for future recommendations on news items. 

\begin{figure*}[!htbp]
\centering
\includegraphics[scale=0.25]{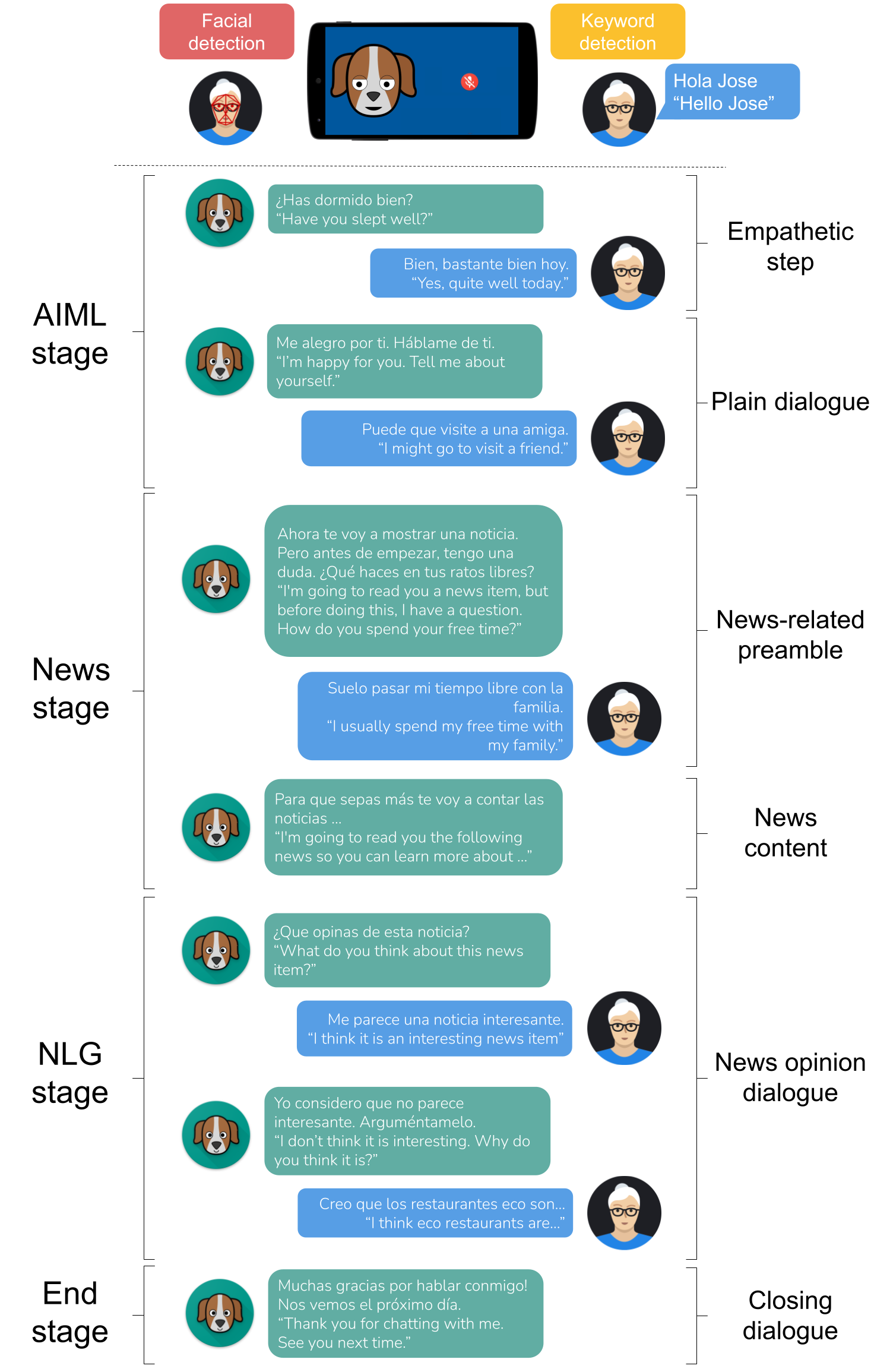}
\caption{\label{flowdiagram} {\sc eber} chatbot flow diagram. Example of a real dialogue.}
\label{dialogoejemplo}
\end{figure*}

Figure \ref{flowdiagram} shows a real example of a whole dialogue between {\sc eber} and a senior taken from the experiments in Section \ref{evaluation}.

\section{Experimental Results and Discussion}
\label{evaluation}

\begin{figure*}[!htbp]
\centering
\includegraphics[scale=0.2]{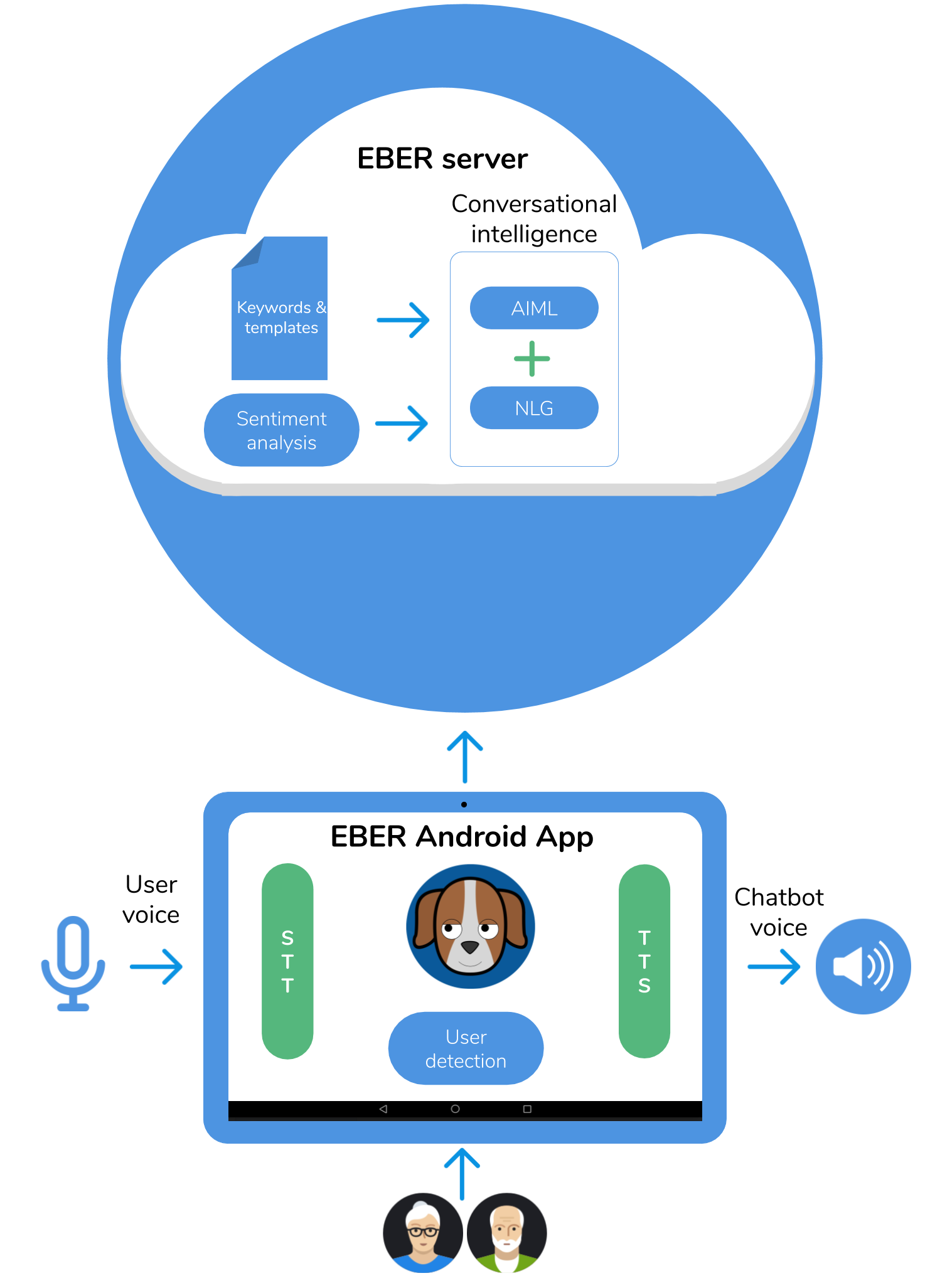}
\caption{\label{server} {\sc eber} testbed.}
\end{figure*}

In the tests an Ubuntu server with 12 cores and 64GB RAM provided remote {\sc sa} and {\sc nlg} capabilities to a user handheld device. The user interface was presented on a Huawei MediaPad T5 tablet with 8 cores, 3GB RAM, 32GB hard disk and Android 8 operating system. Figure \ref{server} shows all the components in the {\sc eber} testbed.

The experiments involved 31 users (20 women and 11 men),
75.5 $\pm$ 6.95 years old (average $\pm$ standard deviation). In the sequel we will call a session a particular newscast and its associated dialogue. The average number of sessions per user was 4.33. Of the 31 participants, 10 had some basic technology skills (such as experience with Google or WhatsApp), 8 had hearing problems, 23 were highly focused during the experiments and 5 felt awkward or stressed during the interactions.

After the dialogue, users were asked if they felt confused, stressed or focused (yes/no in all cases). After each independent session, the users were asked which keywords they would have used to search for similar news in a news repository (logically, users with technology skills immediately identified this exercise as a Google search). When the experiment concluded, users rated their experience, in terms of satisfaction, amazement and chatbot-human likeliness on a 5-level Likert scale.

\subsection{User Satisfaction and Behaviour}

User satisfaction with the service was close to 4 and the perception of chatbot-human likeliness was close to 3 on average. This indicates that even though {\sc ai} technology is not yet mature enough to mimic a human conversation, {\sc eber} was entertaining to the users during the tests (in this regard, 4 users stated that they were amazed by the capability of the chatbot to adapt itself to the user mood. Table \ref{usersdata} presents the responses of these users about empathy). This encouraged us to consider the ``intelligent radio" approach realistic and potentially useful. However, to further validate these results we studied the behaviours of the users during the tests to check if they were coherent with the demographics of the sample and rule out user bias from the experiment setup. 

In this study we checked pairwise correlations between different behavioural variables using the Pearson correlation coefficient \cite{Zhou2016} (see Expression (\ref{pearson}), where $x$ and $y$ are two generic variables). This metric varies between -1 and 1 and is closer to -1 if the relationship is inverse and closer to 1 if it is direct. We considered that two variables were possibly related for absolute correlations around 0.4 or above, since the goal was to obtain preliminary but valuable insights.

\begin{equation}
r_{xy} = \frac{\sum (x_i - \overline{x}) (y_i - \overline{y})}{\sqrt{\sum (x_i - \overline{x})^2} \sqrt{\sum (y_i - \overline{y})^2}}
\label{pearson}
\end{equation}

\begin{figure*}[!htbp]
\centering
\includegraphics[scale=0.5]{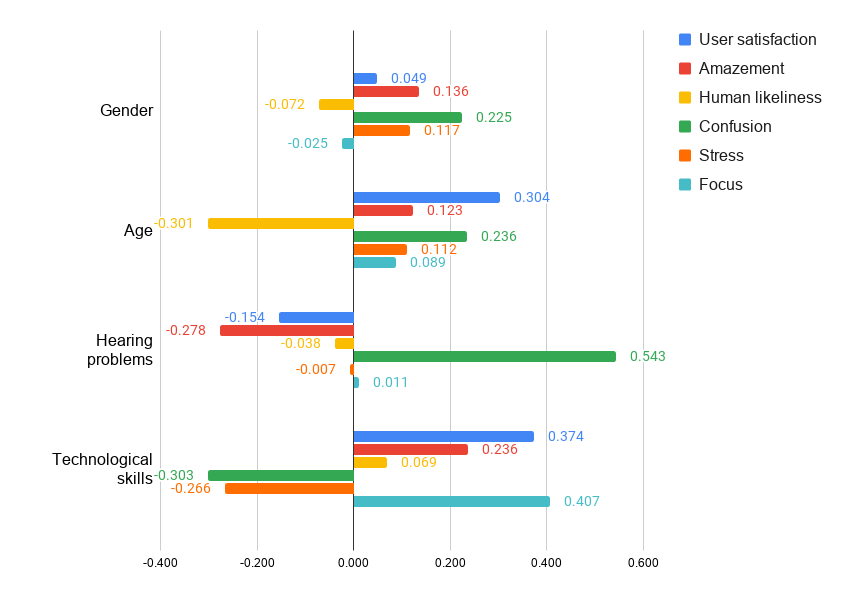}
\caption{\label{figure:resultsDemo} Correlations between behavioural variables and demographics of the sample.}
\end{figure*}

Figure \ref{figure:resultsDemo} shows the correlations between the behavioural variables in the study (user focus, stress and confusion; chatbot-human likeliness; level of amazement with the service; and user satisfaction) and the demographics of the sample. Note that gender was uncorrelated with behaviour. Age only seemed to be correlated with opinion on chatbot-human likeness and user satisfaction: the younger people in the sample were more open to technological novelty while the older people found the system more entertaining (which is especially interesting to us). 

As expected, people with hearing impairments were more confused, less surprised and less satisfied. The level of technology skills had a productive relationship with most of the behavioural variables (the higher that level, the more the focus, amazement and satisfaction and the less stress and confusion). However, it had no clear relationship with opinions on chatbot-human likeness, which suggests that most users were objective regarding this aspect.

\begin{table*}[ht!]
\centering
\small
\caption{\label{usersdata} User answers about empathy of the {\sc eber} chatbot.}
\begin{tabular}{|l|}
\hline
\begin{tabular}[c]{@{}p{12cm}}{}
\hspace{0.5cm} User ID: 3\\

\hspace{1.5cm} Is there anything that you found especially surprising or useful while using the bot?\\
\hspace{1.5cm} {\it The understanding of the bot.}\\\\

\hspace{0.5cm} User ID: 5\\

\hspace{1.5cm} Is there anything that you found especially surprising or useful while using the bot?\\
\hspace{1.5cm} {\it The system moves its eyes and mouth to show a frame of mind.}\\\\

\hspace{0.5cm} User ID: 16\\

\hspace{1.5cm} Is there anything that you found especially surprising or useful while using the bot?\\
\hspace{1.5cm} {\it The systems detects the mood of my opinions.}\\\\

\hspace{0.5cm} User ID: 17\\

\hspace{1.5cm} Is there anything that you found especially surprising or useful while using the bot?\\
\hspace{1.5cm} {\it The chatbot understands me and adapts itself to my answers.}\\

\end{tabular} \\ \hline
\end{tabular}
\end{table*}

Summing up, the most important insight from these measurements was that user satisfaction scores were coherent with expected behaviour, thus supporting their validity. 

\subsection{Explicit Versus Implicit Content Description}
\label{abstraction_measurement}

We measured the ability of the users to describe the content they were presented with by calculating the Normalised Google distance ({\sc ngd}) \cite{Cilibrasi2007} between news metadata and the words the users explicitly chose to search for news when asked to do so. In this regard, some people tended to describe news with lengthy clauses, probably because it was difficult for them to summarise concepts with specific words \cite{Kemper2010}. In such cases we segmented clauses into their word constituents. We discarded stopwords and applied lemmatisation procedures, and the resulting words were taken as separate search terms. For example, the four keywords `good', `beautiful', `cheap' and `restaurants' would be extracted from `good, beautiful and cheap restaurants'. 

The differences among users were notable. Focused users with technology skills selected effective words such as `social services, Spain, Galicia' for news about nursing homes in Galicia, whereas stressed or confused users provided vague terms such as `walk carefully, elderly person, illnesses, too old'. Therefore, we divided the sample into three groups. The first group contained the seven users that were focused and relaxed, not confused, and had good hearing, and the second and third groups, with eleven and thirteen users respectively, everybody else. The tests with the second group, which had similar demographics as the other two (age statistics of 76 $\pm$ 6.75 years in the second group versus 75.23 $\pm$ 7.07 years in the other two) were performed several months later, the main difference being that the users in the third group had worse technological skills.

Table \ref{ngds} shows the {\sc ngd} values for user-provided explicit keywords by group. The gold standard was given by young researchers in our group with high web search skills (average {\sc ngd} of 0.47 for the same news). The average {\sc ngd} was 0.56 (only 0.09 above the gold standard) in the first user group. The technological skills of the users in the second group mitigated discomfort, and their average {\sc ngd} was only a bit worse (0.61 versus 0.56). Of the 13 users in the third group, 4 had very high {\sc ngd} values. Even after discarding the terms that were not directly related to the news' topic, the average {\sc ngd} was still high, of 0.84. As in the previous section, we checked the coherence of these results by analysing correlations. In particular, we found positive correlations between {\sc ngd} values and stress and focus levels. More specifically, stressed and focused users had respective correlations of 0.58 and -0.44. 

\begin{table}[!htbp]
\centering
\caption{\label{ngds}{\sc ngd} values, user-provided explicit keywords.}
\begin{tabular}{cccc}
\hline
& \bf Group 1 & \bf Group 2 & \bf Group 3\\\toprule
& 0.61 & 0.64 & 0.54\\
& 0.50 & 0.59 & 0.31 \\
& 0.55 & 0.60 & 0.74 \\
& 0.56 & 0.60 & 2.73 \\
& 0.58 & 0.69 & 0.56 \\
& 0.58 & 0.62 & 2.23 \\
& 0.52 & 0.54 & 0.85 \\
& & 0.62 & 0.57 \\
& & 0.63 & 0.45 \\
& & 0.59 & 0.48 \\
& & 0.59 & 0.49 \\
& & & 0.48\\
& & & 0.49 \\\hline
\bf AVG & \bf 0.56 & \bf 0.61 & \bf 0.84 \\\bottomrule
\end{tabular}
\end{table}

Finally, we evaluated the knowledge about the news in the dialogue between the chatbot and the users (see example in Figure \ref{dialogoejemplo}). For this purpose we identified the semantic category of the words in the news (excluding stopwords) using the Multilingual Central Repository ({\sc mcr}) tool\footnote{Available at {\tt http://www.talp.upc.edu/demo-detail/440
/MRC}, March 2021.} \cite{Gonzalez-Agirre2012}. Then we extracted from the dialogues the words with the highest semantic similarity with the content of the news. We considered there was a match between the words provided by the users and the words in the news when their {\sc mcr} semantic categories were the same. Next, we built a new set with all the semantically related words (if there was no semantic match, we just employed the user's response). 
The {\sc ngd} was recomputed with this set (Table \ref{correlations2}). Table \ref{correlations3} summarises the results of both experiments. Compared to the gold standard, the average {\sc ngd} decreased from 179\% to 136\% (-0.2), 130\% to 121\% (-0.04) and 119\% to 109\% (-0.05) in groups 3, 2 and 1, respectively.
\begin{table}[!htbp]
\centering
\caption{\label{correlations2}{\sc ngd} values, by adding keywords from semantic knowledge.}
\begin{tabular}{cccc}
\hline
& \bf Group 1 & \bf Group 2 & \bf Group 3\\\toprule
& 0.55 & 0.72 & 0.51\\
& 0.46 & 0.69 & 0.47\\
& 0.53 & 0.60 & 0.57\\
& 0.47 & 0.63 & 2.00\\
& 0.49 & 0.52 & 0.52\\
& 0.48 & 0.55 & 0.47\\
& 0.58 & 0.49 & 0.75\\
& & 0.56 & 0.52\\
& & 0.46 & 0.47\\
& & 0.54 & 0.58\\
& & 0.52 & 0.53\\
& & & 0.55 \\
& & & 0.44 \\\hline
\bf AVG & \bf 0.51 & \bf 0.57 & \bf 0.64\\\bottomrule
\end{tabular}
\end{table}

\begin{table}[!htbp]
\centering
\caption{\label{correlations3}Summary of average {\sc ngd} values: Gold Standard (\sc{gs}), group 1, group 2 and group 3.}
\begin{tabular}{ccccc}
\hline
&\bf GS & \bf Group 1 & \bf Group 2 & \bf Group 3\\\toprule
\begin{tabular}[c]{@{}p{2.5cm}@{}} Explicit keywords \end{tabular} & 0.47 & 0.56 & 0.61 & 0.84\\
\begin{tabular}[c]{@{}p{2.5cm}@{}} Explicit keywords $+$ semantic knowledge \end{tabular} & - & 0.51 & 0.57 & 0.64\\\hline
\end{tabular}
\end{table}

We remark that the results were consistent over time: even though the tests with user group 2 (with similar demographics) were performed much later, they also reflected a systematic improvement in news abstraction capabilities by automatically adding knowledge. In any case, the benefits of the system were more evident in users without technological skills who experienced some discomfort.

We have therefore not only proved that a specialised chatbot can engage elderly people as an ``intelligent newscaster radio", but also that it can extract keywords that help to personalise subsequent news searches if the user expresses positive sentiments about previous news.

\subsection{User Benefits}
\label{using}

From a holistic perspective, the system may open a world of digital information to the users through a familiar channel (background voice or ``radio") with a friendly interface (voice chatbot), and it may also provide value as a tool to measure the evolution of their abstraction capabilities (and thus, their cognitive abilities). It can also enable non-intrusive monitoring from families and caregivers.

During the experiments, the users were also asked about the aspects they liked or disliked. Remarkably, in addition to human-machine communication features (visual indications of user's turn to speak, chatbot empathy and front-end avatar), the newscaster functionality of the system was praised. Regarding the aspects the users disliked, confused users were particularly baffled by chatbot interruptions when they paused for too long. This revealed that we need to develop better ways to detect when a user has finished his/her turn. 

Overall, given the high user satisfaction scores, we concluded that the ``intelligent radio" approach has the potential to help elderly users by improving their abstraction capabilities, which narrows the digital gap.

\section{Conclusions}
\label{conclusions}

In this work we have presented the {\sc eber} intelligent chatbot. To the best of our knowledge this is the first system to combine {\sc aiml}, {\sc nlg} and {\sc sa} to generate short coherent contextualised dialogues as connectors between newscasts. Thanks to this combination, {\sc eber} behaves realistically as an ``intelligent radio" for entertaining elderly people.

In the tests, 80\% of the users gave the system a satisfaction score of 4 out of 5. The analysis of correlations between behavioural variables and sample demographics was coherent, further supporting the validity of the user satisfaction scores. By automatically extracting knowledge from connecting dialogues with positive sentiment, the system improves content characterisation, even for unfocused, stressed or confused people.

In forthcoming work, we will explore how to improve turn detection to avoid speech interruptions and add advanced personalisation features with better directed dialogue stages. Moreover, we will consider an experimental assessment of loneliness and study how the quality of users' responses (regarding user words that complement news keywords for subsequent searches) evolves in time, by paying special attention to differences in speech related to age. We theorise that quality will not worsen thanks to the positive reinforcement users experience when new newscasts match their interests.

It may also be interesting to study the effect of other forms of language semantics such as sarcasm \cite{Jun2016}.

\bibliography{mendeley.bib}{}
\bibliographystyle{IEEEtran}

\EOD

\end{document}